\documentclass{article}
\usepackage{microtype,multirow,hyphenat,hyperref,amsfonts,amsmath,amsthm,bm,graphicx,times}
\usepackage[accepted]{icml2014}
\icmltitlerunning{Randomized Nonlinear Component Analysis}

\newcommand{\reals}{\mathbb{R}}
\newcommand{\ip}[2]{\langle {#1},\, {#2} \rangle}
\newcommand{\E}{\mathbb{E}}
\newcommand{\norm}[1]{\left\|#1\right\|}
\newcommand{\nlsum}{\sum\nolimits}

\newtheorem{theorem}{Theorem}

\begin{document} 
\twocolumn[
  \icmltitle{Randomized Nonlinear Component Analysis}
  \icmlauthor{David Lopez-Paz}{dlopez@tue.mpg.de}\\
  {Max-Planck-Institute for Intelligent Systems, University of Cambridge}\vskip 0.2 cm
  \icmlauthor{Suvrit Sra}{suvrit@tue.mpg.de}\\
  {Max-Planck-Institute for Intelligent Systems, Carnegie Mellon University}\vskip 0.2cm
  \icmlauthor{Alexander J. Smola}{alex@smola.org}\\
  {Carnegie Mellon University, Google Research}\vskip 0.2cm
  \icmlauthor{Zoubin Ghahramani}{zoubin@eng.cam.ac.uk}\\
  {University of Cambridge}\vskip 0.2cm
  \icmlauthor{Bernhard Sch\"olkopf}{bs@tue.mpg.de}\\
  {Max-Planck-Institute for Intelligent Systems}
  \icmlkeywords{component analysis, randomized methods, lupi, autoencoders}
  \vskip 0.3in
]

\begin{abstract}
  Classical methods such as Principal Component Analysis (PCA) and Canonical
  Correlation Analysis (CCA) are ubiquitous in statistics.  However, these
  techniques are only able to reveal linear relationships in data. Although
  nonlinear variants of PCA and CCA have been proposed, these are
  computationally prohibitive in the large scale. 

  In a separate strand of recent research, randomized methods have been
  proposed to construct features that help reveal nonlinear patterns in data.
  For basic tasks such as regression or classification, random features exhibit
  little or no loss in performance, while achieving drastic savings in
  computational requirements.

  In this paper we leverage randomness to design scalable new variants of
  nonlinear PCA and CCA; our ideas extend to key multivariate analysis
  tools such as spectral clustering or LDA. We demonstrate our algorithms
  through experiments on real-world data, on which we compare against the
  state-of-the-art. A simple R implementation of the presented algorithms is
  provided.
\end{abstract} 

\section{Introduction}
  Principal Component Analysis \citep{Pearson01} and Canonical Correlation
  Analysis \citep{Hotelling36} are two of the most popular multivariate analysis
  methods. They have played a crucial role in a vast array of applications
  since their conception a century ago.

  Principal Component Analysis (PCA) rotates a collection of correlated
  variables into their uncorrelated principal components (also known as factors
  or latent variables). Principal components owe their name to the following
  property: the first principal component captures the maximum amount of
  variance in the data; successive components account for the maximum amount of
  remaining variance in dimensions orthogonal to the preceding ones. PCA is
  commonly used for dimensionality reduction, assuming that core properties of
  a high-dimensional sample are largely captured by a small number of principal
  components.

  Canonical Correlation Analysis (CCA) computes linear transformations of a pair of
  random variables such that their projections are maximally correlated.
  Analogous to principal components, the projections of the pair of random
  variables are mutually orthogonal and ordered by their amount of explained
  cross-correlation. CCA is widely used to learn from multiple modalities of
  data \citep{Kakade07}, an ability particularly useful when some of the
  modalities are only available at training time but keeping information about
  them at testing time is beneficial \citep{Chaudhuri09,Vapnik09}.
  
  The applications of PCA and CCA are ubiquitous. Some examples are feature
  extraction, time-series prediction, finance, medicine, meteorology,
  chemometrics, biology, neurology, natural language processing, speech
  recognition, computer vision or multimodal signal processing
  \citep{Jolliffe02}. 
  
  Despite their success, an impediment of PCA and CCA for modern data analysis
  is that both reveal only linear relationships between the variables under
  study. To overcome this limitation, several nonlinear extensions have been
  proposed for both PCA and CCA. For PCA, these include Kernel Principal
  Component Analysis or KPCA \citep{Schoelkopf99} and Autoencoder Neural
  Networks \citep{Baldi89,Hinton06}. For CCA, common extensions are Kernel
  Canonical Correlation Analysis or KCCA \citep{laiFy00,Bach02} and Deep
  Canonical Correlation Analysis \citep{Galen13}. However, these solutions tend
  to have rather high computational complexity (often cubic in the sample
  size), are difficult to parallelize or are not accompanied by
  theoretical guarantees. 

  In a separate strand of recent research, randomized strategies have been
  introduced to construct features that can help reveal nonlinear patterns
  in data when used in conjunction with linear algorithms \citep{Rahimi08,Le13}.
  For basic tasks such as regression or classification, using nonlinear random
  features incurs little or no loss in performance compared with exact kernel
  methods, while achieving drastic savings in computational complexity (from
  cubic to linear in the sample size). Furthermore, random features are
    amenable to simple implementation and clean theoretical analysis.  
    
  The main contribution of this paper is to lay the foundations for nonlinear,
  randomized variants of PCA and CCA. Therefore, we dedicate attention to
  studying the spectral properties of low-rank kernel matrices constructed as
  sums of random feature dot-products.  Our analysis relies on the recently
  developed matrix Bernstein inequality \citep{Tropp14}. With little additional
  effort, our analysis extends to other popular multivariate analysis tools
  such as linear discriminant analysis, spectral clustering and the randomized
  dependence coefficient.

  We demonstrate the effectiveness of the proposed randomized methods by
  experimenting with several real-world data and comparing against the
  state-of-the-art Deep Canonical Correlation Analysis \citep{Galen13}.
  As a novel application of the presented methods, we derive an algorithm to
  learn using privileged information \citep{Vapnik09} and a scalable strategy
  to train nonlinear autoencoder neural networks.  Additional numerical
  simulations are provided to validate the tightness of the concentration
  bounds derived in our theoretical analysis. Lastly, the presented methods are
  very simple to implement; we provide R source code at:
  \begin{center}
    {\footnotesize\url{http://lopezpaz.org/code/rca.r}}
  \end{center}

  \subsection{Related Work}
  There has been a recent stream of research in kernel approximations via
  randomized feature maps since the seminal work of \citet{Rahimi08}. For
  instance, their extensions to dot-product kernels \citep{kar2012random} and
  polynomial kernels \citep{hamid2013compact}; the development of advanced
  sampling techniques using Quasi-Monte-Carlo methods \citep{yang14} or their
  accelerated computation via fast Walsh-Hadamard transforms \citep{Le13}.

  The use of randomized techniques for kernelized component analysis methods
  dates back to \citep{achlio02}, where three kernel sub-sampling strategies
  were suggested to speed up KPCA.  On the other hand, \citep{Avron13} made use
  of randomized Walsh-Hadamard transforms to adapt linear CCA to large-scale
  datasets. The use of non-linear random features is more scarce and has only
  appeared twice in previous literature. First, \citet{Lopez-Paz13} defined the
  dependence statistic RDC as the largest canonical correlation between two
  sets of copula random projections.  Second, \citet{McWilliams13} used the
  Nystr\"om method to define a randomized feature map and performed CCA to
  achieve state-of-the-art semi-supervised learning.
  
  \section{Random Nonlinear Features}
  \label{sec:randomized}
  We start our presentation by recalling a few key aspects of nonlinear random
  features.

  Consider the class $\mathcal{F}_p$ of functions whose weights decay faster than some sampling
  distribution $p$; formally:
  \begin{equation}\label{eq:funclass}
    \mathcal{F}_p := \left\lbrace f(\bm x) = \int_{\mathbb{R}^{d}} \alpha(\bm w)
    \phi(\bm x^T\bm w) \mathrm{d}\bm w : |\alpha(\bm w)|\leq Cp(\bm w)
    \right\rbrace,
  \end{equation}
  Here, $\alpha : \reals^d \to \reals$ is a
  nonlinear map of ``weights'',  while $\phi : \mathbb{R} \rightarrow
  \mathbb{R}$ is a nonlinear map that satisfies $|\phi(z)| \leq 1$; $\bm x, \bm
  w$ are vectors in $\mathbb{R}^{d}$, $p(\bm w)$ is a probability density of
  the parameter vectors $\bm w$ and $C$ is a regularizing constant. More
  simply, we may consider the finite version of $f$:
  \begin{equation}\label{eq:finite}
      {f_m}(\bm x) := \sum\nolimits_{i=1}^m \alpha_i \phi(\bm{w}_i^T\bm x).
    \end{equation}
  Kernel machines, Gaussian processes, AdaBoost, and neural networks are models
  that fit within this function class.

  Let $\mathcal{D} = \{(\bm x_i, y_i)\}_{i=1}^n \subset \mathbb{R}^d \times
  \mathbb{R}$ be a finite sample of input-output pairs drawn from a
  distribution $Q(X,Y)$.  We seek to approximate a function $f$ in class
  $\mathcal{F}_p$ by minimizing the empirical risk (over dataset $\cal D$)
  \begin{equation}
    \label{eq:emp}
    R_{\text{emp}}(f) := \frac{1}{m}\sum\nolimits_{i=1}^m c(f_m(\bm{x}_i),y_i),
  \end{equation}
  for a suitable loss function $c(\hat{y},y)$ that penalizes departure of
    $f_m(\bm x)$ from the true label $y$; for us, the least-squares loss
    $c(\hat{y},y)=(\hat{y}-y)^2$ will be most convenient. 

  Jointly optimizing~\eqref{eq:emp} over $(\bm
  \alpha,\bm{w}_1,\ldots,\bm{w}_m)$ used in defining $f_m$,   is a daunting
  task given the nonlinear nature of $\phi$.  The key insight of
  \citet{Rahimi08} is that we can instead randomly sample the parameters $\bm
  w_i \in \mathbb{R}^{d}$ from a data-independent distribution $p(\bm w)$ and
  construct an $m$-dimensional randomized feature map $\bm z(\bm X)$ for the
  input data $\bm X \in \mathbb{R}^{n \times d}$ that obeys the following
  structure: 
  \begin{align}\label{eq:theaug}
    &\bm w_1, \ldots, \bm w_m \sim p(\bm w),\nonumber\\
    \bm z_i      &:= [\cos(\bm w_i^T\bm x_1 + b_i), \ldots, \cos(\bm w_i^T\bm
    x_n + b_i)]
    \in \mathbb{R}^{n},\nonumber\\
    \bm z(\bm X) &:=  \left[ \bm z_1 \cdots \bm z_m \right] \in \mathbb{R}^{n
    \times m}.
  \end{align}
  Using the (precomputed) nonlinear random features $\bm z(\bm X)$ ultimately
  transforms the nonconvex optimization of (\ref{eq:emp}) into a least-squares
  problem of the form
  \begin{equation}
    \label{eq:lsq}
    \min_{\bm \alpha \in \mathbb{R}^d} \nolimits \left\|\bm y-\bm 
    z(\bm X)\bm \alpha\right\|^2_2,\quad \text{s.t.}\ \norm{\bm
    \alpha}_{\infty} \le C.
  \end{equation}
  This form remarkably simplifies computation (in practice, we solve a
  regularized version of it), while incurring only a bounded error.
  Theorem~\ref{thm:rahimi} formalizes this claim.
  
  \begin{theorem}\label{thm:rahimi}
  \citep{Rahimi08} Let $\mathcal{F}_p$ be defined as in (\ref{eq:funclass}).
  Draw $\mathcal{D} \sim Q(\bm X,Y)$. Construct $\bm z(\cdot)$ as in
  (\ref{eq:theaug}). Let $c : \mathbb{R}^2 \rightarrow \mathbb{R}_+$ be a
  loss-function $L$-Lipschitz in its first argument.  Then, for any $\delta >
  0$, with probability at least $1-2\delta$ there exist some $\bm \alpha =
  (\alpha_1, \ldots, \alpha_m)$ such that
  \begin{align*}
   &\mathbb{E}_Q\left[c\left(\bm z(\bm x)\bm \alpha,y\right)\right]-\\
   \min_{f\in \mathcal{F}_p}&\E_Q[c(f(\bm x),y)]
   \leq O\left(\left( \tfrac{LC}{\sqrt{n}} +
   \tfrac{LC}{\sqrt{m}} \right) \sqrt{\log \tfrac{1}{\delta}}\right).
  \end{align*}
  \end{theorem}

  Solving~\eqref{eq:lsq} takes $O(ndm+m^2n)$ operations, while testing $t$
  points on the fitted model takes $O(tdm)$ operations.  Recent techniques that
  use subsampled Hadamard randomized transforms \citep{Le13} allow faster computation
  of the random features, yielding $O(n\log(d)m+m^2n)$ operations
  to solve~\eqref{eq:lsq} and $O(t\log(d)m)$ to test $t$ new points.

  It is of special interest that randomized algorithms are in many cases more
  robust than their deterministic analogues \citep{Mahoney11} because of the
  \emph{implicit regularization} induced by randomness.

  \subsection{Random Features, Nystr\"om and Kernel Matrices}\label{sec:bochner}
  Bochner's theorem helps connect shift-invariant kernels \citep{Scholkopf01}
  and random nonlinear features. Let $k(\bm x,\bm y)$ be a real
  valued, normalized ($k(\bm x, \bm y) \leq 1$), shift-invariant kernel on
  $\reals^d \times \reals^d$.  Then, 
  \begin{align*}
    k(\bm x, \bm y) &= \int_{\reals^d} p(\bm w) e^{-\mathrm{j}\bm w^T(\bm x-\bm
    y)}\mathrm{d}\bm w\\
                     &\approx \sum\nolimits_{i=1}^m \tfrac{1}{m}
                     e^{-\mathrm{j}\bm w_i^T\bm x}e^{\mathrm{j}\bm w_i^T\bm
                     y}\\
                     &= \sum\nolimits_{i=1}^m \tfrac{1}{m} \cos(\bm w_i^T\bm
                     x+ b_i)\cos (\bm w_i^T\bm y+ b_i)\\\
                     &= \ip{\tfrac{1}{\sqrt{m}}\bm z(\bm
                     x)}{\tfrac{1}{\sqrt{m}}\bm z(\bm y)},
  \end{align*}
  where $p(\bm w)$ is set to be the inverse Fourier transform of $k$ and $b_i
  \sim \mathcal{U}(0,2\pi)$ \citep{Rahimi08}---e.g., the Gaussian kernel $k(\bm
  x,\bm y)=\exp(-s\|\bm x-\bm y\|_2^2)$ can be approximated  using $\bm w_i
  \sim \mathcal{N}(\bm 0, 2s\bm I)$.

  Let $\bm K \in \mathbb{R}^{n\times n}$ be the kernel matrix of some data $\bm
  X \in \mathbb{R}^{n\times d}$, i.e., $\bm K_{ij} = k(\bm x_i, \bm x_j)$ .
  When approximating the kernel $k$ using $m$ random Fourier features, we may
  as well approximate the kernel matrix $\bm K \approx
  \hat{\bm K}$, where
  \begin{equation}\label{eq:kmats}
    \hat{\bm K} := \frac{1}{m}\bm z(\bm X)\bm z(\bm X)^T =
    \frac{1}{m} \sum_{i=1}^m \bm z_i\bm z_i^T = \sum_{i=1}^m \bm \hat{\bm
    K}^{(i)}.
  \end{equation}
  The focus of this paper is on building scalable kernel component analysis
  methods which not only exploit these approximations but are also accompanied
  by theoretical guarantees.

  Importantly, our analysis extends straight-forwardly to features constructed
  using the Nystr\"om method \citep{Seeger01} when its basis are bounded and sampled at
  random. Recent theoretical and empirical evidence suggest the superiority of
  the Nystr\"om method when compared to the aforementioned Fourier features
  \citep{Yang12}. However, we did not experience large differences
  (Section \ref{sec:experiments}), while random Fourier features are faster to
  compute and do not need to be stored at test time \citep{Le13}.

  \section{Principal Component Analysis (PCA)} \label{sec:pca}
  Principal Component Analysis or PCA \citep{Pearson01,Jolliffe02} is the
  orthogonal transformation of a set of $n$ observations of $d$ correlated
  variables $\bm X\in\mathbb{R}^{n \times d}$ into a set of $n$ observations of
  $d$ uncorrelated \emph{principal components}.

  For a centered data matrix (zero mean columns) $\bm X$, PCA requires
  computing the (full) singular value decomposition
  \begin{equation*}
    \bm X = \bm U\bm \Sigma \bm F^T,
  \end{equation*}
  where $\bm \Sigma$ is a diagonal matrix containing the singular values of
  $\bm X$ in decreasing order. The principal components are computed via the
  {linear transformation} $\bm X\bm F$.

  {Nonlinear} variants of PCA are also known; notably,
  \begin{list}{$\bullet$}{\leftmargin=1em}
    \vspace*{-5pt}
    \setlength{\itemsep}{-1pt}
   \item Kernel PCA \citep{Schoelkopf99} uses the \emph{kernel trick} to embed
   data into a high-dimension Reproducing Kernel Hilbert Space, where regular
   PCA is performed. Computation of the principal components reduces to an
   eigenvalue problem, which takes $O(n^3)$ operations.
   \item Autoencoders \citep{Hinton06} are artificial neural networks configured
   to learn their own input.  They are trained to learn compressed
   representations of data. The transformation computed by a linear autoencoder
   with a bottleneck of size $r < d$ is the projection into the subspace
   spanned by the first $r$ principal components of the training data
   \citep{Baldi89}.
  \end{list}

  \subsection{Randomized Nonlinear PCA (RPCA)}
  We propose RPCA, a nonlinear randomized variant of PCA. We may view RPCA
  as a low-rank approximation of KPCA when the latter is equipped with a
  shift-invariant kernel. RPCA may be thus understood as linear PCA on a
  randomized nonlinear mapping of the data. Schematically,
  \begin{align*}
    \{ \bm F, \bm z(\cdot) \} =: \mathrm{RPCA}(\bm X) \equiv \mathrm{PCA}(\bm z(\bm
    X)) \approx \mathrm{KPCA}(\bm X),
  \end{align*}
  where $\bm F \in \mathbb{R}^{m\times m}$ are the principal component loading
  vectors and $\bm z : \mathbb{R}^{n\times d} \rightarrow \mathbb{R}^{n\times
  m}$ is a random feature map generated as in \eqref{eq:theaug} (typically, $m
  \ll n$).

  The computational complexity is $O(n^3)$ for KPCA, $O(d^2n)$ for PCA and
  $O(m^2n)$ for RPCA.  PCA and RPCA are both linear in the sample size $n$.
  When using nonlinear features as in (\ref{eq:theaug}), PCA loadings are no
  longer linear transformations but approximations of nonlinear functions
  belonging to the function class $\mathcal{F}_p$ described in Section
  \ref{sec:randomized}.
  
  As a consequence of Bochner's theorem (Section \ref{sec:bochner}), the RPCA
  kernel matrix will approximate the one of KPCA as the number of random features $m$ tends to infinity.
  This is because random feature dot-products converge uniformly to the exact
  kernel evaluations in expectation \citep{Rahimi08}.  Since the solution of
  KPCA is the eigensystem of the kernel matrix $\bm K$ for the data matrix $\bm
  X \in \reals^{n\times d}$, one may study how fast the approximation $\hat{\bm
  K}$ made in \eqref{eq:kmats} converges in operator (or spectral) norm to $\bm
  K$ as $m$ grows.
  
  To analyze this convergence we appeal to the recently proposed Matrix
  Bernstein Inequality. In the theorem below and henceforth $\|\bm X\|$ denotes
  the operator norm.
  \begin{theorem}[Matrix Bernstein, \citep{Tropp14}]\label{thm:bernstein} Let
  $\bm Z_1,\ldots \bm Z_m$ be independent $d\times d$ random matrices. Assume
  that $\E[\bm Z_i]=0$ and that $\norm{\bm Z_i} \le R$. Define the variance
  parameter $\sigma^2 := \max\bigl\lbrace \norm{\sum\nolimits_i
  \E[\bm{Z}_i^T\bm{Z}_i]}, \norm{\sum\nolimits_i
  \E[\bm{Z}_i\bm{Z}_i^T]}\bigr\rbrace$.  Then, for all $t \ge 0$,
  \begin{equation*}
    \mathbb{P}\left(\norm{\sum\nolimits_i\bm Z_i} \geq t\right)
    \leq 2d \cdot \exp\Bigl\lbrace\frac{-t^2}{3\sigma^2+2Rt}\Bigr\rbrace.
  \end{equation*}
  Furthermore,
  \begin{equation*}
    \mathbb{E}\norm{\nlsum_i\bm Z_i} \leq   \sqrt{3\sigma^2\log(2d)}+ R\log(2d).
  \end{equation*}
  \end{theorem}
  The convergence rate of RPCA to its exact kernel counterpart KPCA is
  expressed by the following  theorem, which actually invokes the Hermitian
  matrix version of Theorem~\ref{thm:bernstein}, and hence depends on $d$
  instead of $2d$, and uses matrix squares when defining the variance
  $\sigma^2$.
  \begin{theorem}\label{thm:pca}
    Assume access to the data $\bm X \in \mathbb{R}^{n \times d}$ and a
    shift-invariant, even kernel $k$. Construct the
    kernel matrix $\bm K_{ij} := k(\bm x_i, \bm x_j)$ and its approximation
    $\hat{\bm K}$ using $m$ random features as per (\ref{eq:kmats}). Then,
    \begin{equation}\label{eq:pcaconc}
        \mathbb{E} \| {\hat{\bm K}} - \bm K\| \leq \sqrt{\frac{3n^2\log
        n}{m}} + \frac{2n\log n}{m}.
    \end{equation}
    \begin{proof} 
    We follow a derivation similar to~\citet{Tropp12b}. 
    
    Denote by
    \begin{equation*}
      {\hat{\bm K}} := \tfrac{1}{m} \sum\nolimits_{i=1}^m \bm z_i\bm z_i^T =
      \sum\nolimits_{i=1}^m
      {\hat{\bm K}^{(i)}}
    \end{equation*}
    the $n \times n$ sample kernel matrix, and by $\bm {K}$ its population
    counterpart such that $\mathbb{E}[{\hat{\bm K}}] = \bm K$. Note that
    ${\hat{\bm K}}$ is the sum of the $m$ independent matrices ${\hat{\bm
    K}^{(i)}}$, since our random features are sampled i.i.d. and the matrix
    $\bm X$ is defined to be constant.  Consider the individual error
    matrices
    \begin{equation*}
      \bm E = {\hat{\bm K}} - \bm K = \sum\nolimits_{i=1}^m \bm E_i, \quad \bm
      E_i = \frac{1}{m} ({\hat{\bm K}^{(i)}}- \bm K),
    \end{equation*}
    each of which satisfies $\mathbb{E}[\bm E_i] = \bm 0$. Since we are using
    bounded features---see $\bm z(\bm x)$ in \eqref{eq:theaug}---it follows that
    there exists a constant $B$ such that $\|\bm z\|^2 \leq B$. Thus, we see
    that
    \begin{equation*}
      \| \bm E_i \| = \frac{1}{m} \|\bm z_i\bm z_i^T - \mathbb{E}[\bm z\bm
      z^T]\| \leq \frac{1}{m}(\|\bm z_i\|^2 + \mathbb{E}[\|\bm z\|^2]) \leq
      \frac{2B}{m},
    \end{equation*}
    because of the triangle inequality on the norm and Jensen's inequality on
    the expected value.  To bound the variance of $\bm E$, bound  first the
    variance of each of its sumands $\bm E_i$ (noting that
    $\mathbb{E}[\bm{z}_i\bm{z}_i^T]=\bm{K}$):
      \begin{align*}
        \mathbb{E}[\bm E_i^2] &= \frac{1}{m^2} \mathbb{E}\left[(\bm z_i\bm z_i^T
        - \bm K)^2\right] \\
        &=\frac{1}{m^2} \mathbb{E} \left[\|\bm z_i\|^2\bm z_i\bm z_i^T - 
        \bm z_i\bm z_i^T\bm K - \bm{K}\bm{z}_i\bm{z}_i^T
        + \bm K^2\right]\\ 
      &\preceq \frac{1}{m^2} \left[
        B\bm K - 2\bm K^2 + \bm K^2\right] \preceq \frac{B\bm
        K}{m^2}.
      \end{align*}
      Next, taking all summands $\bm E_i$ together we obtain
      \begin{equation*}
        \|\mathbb{E}[\bm E^2]\| \leq \left\|\sum\nolimits_{i=1}^m \mathbb{E}\bm
        E_i^2\right\| \leq \frac{1}{m}{B\|\bm K\|}.
      \end{equation*}
      Where the first inequality follows by Jensen. We can now invoke the
      matrix Bernstein inequality (Theorem \ref{thm:bernstein}) on $\bm E - \E[\bm{E}]$ to
      obtain the bound
      \begin{equation*}
        \mathbb{E} \| {\hat{\bm K}} - \bm K\| \leq \sqrt{\frac{3B\|\bm K\|\log n}{m}} +
        \frac{2B\log n}{m}.
      \end{equation*}
      Observe that random features and kernel evaluations are upper-bounded by
      $1$; thus both $B$ and $\|\bm K\|$ are upper-bounded by $n$, yielding the
      bound~(\ref{eq:pcaconc}).
    \end{proof}
  \end{theorem}
  To obtain a characterization in relative-error, we can divide both sides
  of~\eqref{eq:pcaconc} by $\|\bm K\|$. This results in a bound that depends on
  $n$ logarithmically (since $\|\bm K\| = O(n)$). Moreover, additional
  information may be extracted from the tail-probability version of Theorem
  \label{thm:bernstein}.  Please refer to Section \ref{sec:expbounds} for
  additional discussion on this aspect.

  Before closing this section, we mention a byproduct of our above
  analysis.
  
  \textbf{Extension to Spectral Clustering.}
  Spectral clustering~\citep{Luxburg07}  uses the spectrum of $\bm K$ to perform
  dimensionality reduction before applying $k$-means. Therefore, the analysis
  of RPCA may be easily extended to obtain a randomized and nonlinear variant
  of spectral clustering.

  \section{Canonical Correlation Analysis (CCA)}
  \label{sec:cca}
  
  Canonical Correlation Analysis or CCA \citep{Hotelling36} measures the
  correlation between two multidimensional random variables.  Specifically,
  given two samples $\bm X \in \mathbb{R}^{n\times p}$ and $\bm Y \in
  \mathbb{R}^{n\times q}$, CCA computes a pair of \emph{canonical} bases $\bm F \in
  \mathbb{R}^{p \times r}$ and $\bm G \in \mathbb{R}^{q \times r}$ such that 
  \begin{align*}
    \|&\mathrm{corr}(\bm X\bm F,\bm Y\bm G)-\bm I\|_F \text{ is minimized},\\
      &\mathrm{corr}(  \bm X\bm F,\bm X\bm F) = \bm I,\quad \mathrm{corr}(  \bm
      Y\bm G,\bm Y\bm G) = \bm I,
  \end{align*}
  where $\bm I$ stands for the identity matrix. The correlations between the
  \emph{canonical} variables $\bm X\bm F$ and $\bm Y\bm G$ are referred to as
  \emph{canonical correlations}, and up to $r =\text{max}(\text{rank}(\bm X),
  \text{rank}(\bm Y))$ of them can be computed.  The canonical correlations
  $\rho^2_1, \ldots, \rho^2_r$ and basis vectors $\bm f_1,\ldots, \bm f_r \in
  \mathbb{R}^p$ and $\bm g_1, \ldots, \bm g_r \in \mathbb{R}^q$ form the
  eigensystem of the generalized eigenvalue problem \citep{Bie05}:
  \begin{align*}
    \left(
        \begin{array}{cc}
         \bm 0 &  \bm C_{XY}\\
         \bm C_{YX} & \bm 0
        \end{array}
        \right)
    &\left(
        \begin{array}{c}
         \bm f\\
         \bm g 
        \end{array}
        \right)
    =\\
    \rho^2
    \left(
        \begin{array}{cc}
         \bm C_{XX} + \gamma_x \bm I & \bm 0\\
         \bm 0 & \bm C_{YY} + \gamma_y \bm I
        \end{array}
        \right)
    &\left(
        \begin{array}{c}
         \bm f\\
         \bm g 
        \end{array}
        \right)
    ,\nonumber
  \end{align*}
  where $ \bm C_{XY}$ is the covariance $\text{cov}(\bm X, \bm Y)$, while the
  diagonal terms $\gamma \bm I$ act as regularization.

  In another words, CCA processes two different views of the same data (i.e.,
  speech audio signals and paired speaker video frames) and returns their
  maximally correlated linear transformations. This is particularly useful when
  the two views are available at training time, but only one of them is
  available at test time \citep{Kakade07,Chaudhuri09,Vapnik09}.
  
  Several nonlinear extensions of CCA have been proposed:
  \begin{list}{$\bullet$}{\leftmargin=1em}
    \vspace*{-5pt}
    \setlength{\itemsep}{-1pt}
    \item Kernel Canonical Correlation Analysis or KCCA \citep{laiFy00,Bach02}
    uses the {kernel trick} to derive a nonparametric, nonlinear regularized
    CCA algorithm. Its exact computation takes time $O(n^3)$.
    \item Deep Canonical Correlation Analysis or DCCA \citep{Galen13} feeds the
    pair of input variables through a deep neural network. Transformation
    weights are learnt using gradient descent to maximize the correlation of
    the output mappings.
  \end{list}
  \iffalse
  As it occurred with PCA and Neural Network Autoencoders, there exists a very
  tight relationship between CCA and Heteroencoders Neural Network
  \citep{Roweis99}. The widely used denoising autoencoders \citep{Vincent08} are
  a clever use of an heteroencoder.
  \fi

  \subsection{Randomized Nonlinear CCA (RCCA)}\label{sec:rcca}
  We now propose RCCA, a nonlinear and randomized variant of CCA. As will be
  shown, RCCA is a low-rank approximation of KCCA when the latter is equipped
  with a pair of shift-invariant kernels.  RCCA can be understood as linear CCA
  performed on a pair of randomized nonlinear mappings (see \ref{eq:theaug}):
  $\bm z_x : \mathbb{R}^{n\times p}\rightarrow \mathbb{R}^{n\times m_x}$, $\bm
  z_y : \mathbb{R}^{n\times q} \rightarrow \mathbb{R}^{n\times m_y}$ of the
  data $\bm X \in \mathbb{R}^{n \times p}$ and $\bm Y \in \mathbb{R}^{n \times
  q}$.  Schematically,
  \begin{align*}
    \mathrm{RCCA}(\bm X, \bm Y) :=
    \mathrm{CCA}(\bm z_x(\bm X),\bm z_y(\bm Y))
    \approx \mathrm{KCCA}(\bm X, \bm Y).
  \end{align*}
 
  The computational complexity is $O(n^3)$ for KCCA, $O((p^2+q^2)n)$ for CCA
  and $O((m_x^2+m_y^2)n)$ for RCCA. CCA and RCCA are both linear in the sample
  size $n$.

  When performing RCCA, the basis vectors $\bm f_1,\ldots, \bm f_r \in
  \mathbb{R}^p$ and $\bm g_1, \ldots, \bm g_r \in \mathbb{R}^q$ become the
  basis functions $\bm f_1, \ldots, \bm f_r : \mathbb{R}^p \rightarrow
  \mathbb{R}$ and $\bm g_1, \ldots, \bm g_r : \mathbb{R}^q \rightarrow
  \mathbb{R}$, which approximate functions in the class $\mathcal{F}_p$ defined
  in Section \ref{sec:randomized}.
  
  As with PCA, we are interested in characterizing the convergence rate of RCCA
  to its exact kernel counterpart KCCA as $m_x$ and $m_y$ grow. The solution of
  KCCA is the eigensystem of the matrix $\bm R^{-1} \bm L$, where, 
  \begin{align}
    \bm {R}^{-1} & := \begin{pmatrix} (\bm K_x+\gamma_x \bm I)^{-1} & \bm 0\\
    \bm 0 & (\bm K_y +\gamma_y \bm I)^{-1}\end{pmatrix},\label{eq:matr}\\ \bm L
    & := \begin{pmatrix} \bm 0 & \bm K_y\\ \bm K_x & \bm
    0\end{pmatrix},\label{eq:matl}
  \end{align}
  and $(\gamma_x,\gamma_y)$ are positive regularizers mandatory to avoid
  spurious $\pm 1$ correlations \citep{Bach02}.  Theorem \ref{thm:cca}
  characterizes the convergence rate of RCCA to KCCA. Let $\hat{\bm R}^{-1}$
  and $\hat{\bm L}$ be the approximations to~\eqref{eq:matr}
  and~\eqref{eq:matl} obtained by using $m$ random features; that is
  \begin{align}
    \hat{\bm R}^{-1} & := \begin{pmatrix} (\hat{\bm K}_x+\gamma_x \bm I)^{-1} &
    \bm 0\\ \bm 0 & (\hat{\bm K}_y +\gamma_y \bm I)^{-1}
    \end{pmatrix},\label{eq:matrh}\\ \hat{\bm L} & := \begin{pmatrix} \bm 0 &
    \hat{\bm K}_y \\ \hat{\bm K}_x & \bm 0 \end{pmatrix}\label{eq:matlh}.
  \end{align}
  \begin{theorem}\label{thm:cca}
    Assume access to the datasets $\bm X \in \mathbb{R}^{n \times p}$, $\bm Y
    \in \mathbb{R}^{n\times q}$ and shift-invariant kernels $k_x$, $k_y$.
    Define the kernel matrices $(\bm K_x)_{ij} := k_x(\bm x_i, \bm x_j)$,
    $(\bm K_y)_{ij} := k_y(\bm y_i, \bm y_j)$ and their approximations $\hat{\bm
    K}_x$, $\hat{\bm K}_y$ using $m_x$, $m_y$ random features as in
    (\ref{eq:theaug}), respectively.  Let $\bm L$, $\bm R$, $\hat{\bm L}$,
    $\hat{\bm R}$ be as defined in (\ref{eq:matr}--\ref{eq:matlh}), where
    $\gamma_x, \gamma_y > 0$ are regularization parameters.  Furthermore,
    define $\gamma := \min(\gamma_x,\gamma_y)$, $m := \min(m_x,m_y)$. Then,
    \begin{equation}\label{eq:ccaconc}
    \mathbb{E}\| \hat{\bm R}^{-1} \hat{\bm L} - \bm R^{-1}\bm L\| \leq
    \frac{1}{\gamma}\left(\sqrt{\frac{3n^2\log2n}{m}}+\frac{2n\log 2n}{m}
    \right ).
    \end{equation}
    \begin{proof} As the matrices are block-diagonal, we have
    \begin{align*}
    \mathbb{E}\| \hat{\bm R}^{-1} \hat{\bm L} - &\bm R^{-1}\bm L\|\\
    \leq \max(&\mathbb{E}\|(\hat{\bm K}_x+\gamma_x \bm I)^{-1}\hat{\bm
    K}_y-(\bm K_x+\gamma_x \bm I)^{-1}\bm K_y\|,\\
    &\mathbb{E}\|(\hat{\bm K}_y+\gamma_y \bm I)^{-1}\hat{\bm K}_x-(\bm
    K_y+\gamma_y \bm I)^{-1}\bm K_x\|).
    \end{align*}
    We analyze the first term of the maximum; the latter can be analyzed
    analogously.  Let $\hat{\bm A} := (\hat{\bm K}_x+\gamma_x \bm I)^{-1}$ and
    $\bm A := (\bm K_x+\gamma_x \bm I)^{-1}$. Define the individual error terms
    \begin{equation*}
    \bm E_i = \tfrac{1}{m_y}\bigl(\hat{\bm A}\hat{\bm K}^{(i)}_y-\bm A\bm K_y
    \bigr), \quad \bm E = \nlsum_{i=1}^{m_y} \bm E_i.
    \end{equation*}
    Recall that the $m_x+m_y$ random features are sampled i.i.d.\ and that the
    data matrices $\bm X$, $\bm Y$ are constant.  Therefore, the random
    matrices $\bm \hat{\bm K}_x^{(1)}, \ldots, \bm \hat{\bm K}_x^{(m_x)},\bm
    \hat{\bm K}_y^{(1)}, \ldots, \bm \hat{\bm K}_y^{(m_y)}$ are i.i.d.
    random variables.  Hence, their expectations factorize:
    \begin{equation*}
    \mathbb{E}\,[\bm E_i] = \tfrac{1}{m_y}\,\bigl(\mathbb{E}[\hat{\bm A}]\bm
    K_y - \bm A\bm K_y\bigr),
    \end{equation*}
    where we used $\E[\hat{\bm{K}}_y^{(i)}]=\bm K_y$.  The deviation of the
    individual error matrices from their expectations is
    \begin{equation*}
    \bm Z_i := \bm E_i-\mathbb{E}\,[\bm E_i] = \tfrac{1}{m_y}\bigl(\hat{\bm
    A}\hat{\bm K}^{(i)}_y-\mathbb{E}[\hat{\bm A}]\bm K_y\bigr),
    \end{equation*}
    and the norm of this deviation is bounded as
    \begin{equation*}
    \|\bm Z_i\| =\frac{1}{m_y}\|\hat{\bm A}\hat{\bm
    K}^{(i)}_y-\mathbb{E}[\hat{\bm A}]\bm K_y\| \leq \frac{2B}{m_y\gamma_x} =:
    R.
    \end{equation*}
    The inequality follows by applying H\"older twice after using the triangle
    inequality. We now turn to the issue of computing the variance, which is
    defined as
    \begin{equation*}
      \sigma^2 := \max\left\lbrace\norm{\nlsum_{i=1}^{m_y}\E[\bm Z_i\bm
      Z_i^T]},\norm{\nlsum_{i=1}^{m_y}\E[\bm Z_i^T \bm Z_i]}\right\rbrace.
    \end{equation*}
    Consider first second argument of the maximum above, for which we expand an
    individual term in the summand:
    \begin{align*}
      \bm Z_i^T \bm Z_i &= \frac{1}{m_y^2}\Big(\hat{\bm K}_y^{(i)}\hat{\bm
      A}^2\hat{\bm K}_y^{(i)} + \bm K_y\E[\hat{\bm A}]^2\bm K_y \\
      &- \hat{\bm K}_y^{(i)}\hat{\bm A}\E[\hat{\bm A}]\bm K_y - \E[\hat{\bm
      A}]\bm K_y\hat{\bm A}\hat{\bm K}_y^{(i)}\Big).
    \end{align*}
    Taking expectations we see that
    \begin{align*}
        \E[\bm Z_i^T \bm Z_i] &= \frac{1}{m_y^2}\left( \E[\hat{\bm
        K}_y^{(i)}\hat{\bm A}^2\hat{\bm K}_y^{(i)}] - \bm K_y\E[\hat{\bm
        A}]^2\bm K_y\right)\\
        &\preceq \frac{1}{m_y^2}\E[\hat{\bm K}_y^{(i)}\hat{\bm A}^2\hat{\bm K}_y^{(i)}],
    \end{align*}
    where the inequality follows as $\bm K_y\E[\hat{\bm A}]^2\bm K_y\succeq 0$.
    Taking norms and invoking Jensen's inequality we then obtain
    \begin{equation*}
        \norm{\E[\bm Z_i^T\bm Z_i]} \le \frac{B\|\bm K_y\|}{m^2\gamma^2}.
    \end{equation*}
    A similar argument shows that
    \begin{equation*}
      \E[\bm Z_i \bm Z_i^T]  \preceq \frac{1}{m_y^2}\E[\hat{\bm A}(\hat{\bm
      K}_y^{(i)})^2\hat{\bm A}] \Rightarrow
      \|\E[\bm Z_i \bm Z_i^T]\| \le \frac{B\|\bm K_y\|}{m^2\gamma^2}.
    \end{equation*}
    An invocation of Jensen on the definition of $\sigma^2$ along with the two
    bounds above yields the worst-case estimate
    \begin{equation*}
      \sigma^2 \le  \frac{B\|\bm K_y\|}{m_y\gamma^2}.
    \end{equation*}
    We may now appeal to the matrix Bernstein inequality
    (Theorem \ref{thm:bernstein}) to obtain the bound
    \begin{align*}
    \mathbb{E}\,\|(\hat{\bm K}_x+\gamma_x \bm I)^{-1}\hat{\bm K}_y-({\bm
    K}_x+\gamma_x \bm I)^{-1}{\bm K}_y\| \leq\\
    \frac{1}{\gamma_x}\left(\sqrt{\frac{3n^2\log 2n}{m_y}}+\frac{2n\log 2n}{m_y}
    \right ).
    \end{align*}
    The result follows by analogously bounding
    $\mathbb{E}\,\|(\hat{\bm K}_y+\gamma_y \bm I)^{-1}\hat{\bm K}_x-({\bm
    K}_y+\gamma_y \bm I)^{-1}{\bm K}_x\|$ and taking maxima.
    \end{proof}
  \end{theorem}

  Before concluding this section, we briefly comment on two easy extensions of
  our above result. 

  \paragraph{Extension to Linear Discriminant Analysis.}
  Linear Discriminant Analysis (LDA) seeks a linear combination of the features
  of the data $\bm X \in \mathbb{R}^{n\times d}$ such that the samples become
  maximally separable with respect to a paired labeling $\bm y$ with $y_i \in
  \{1,\ldots,c\}$. LDA can be solved by $\mathrm{CCA}(\bm X,\bm T)$, where $\bm
  T_{ij} = \mathbb{I}\{y_i = j\}$ \citep{Bie05}.  Therefore, a similar analysis
  to the one of RCCA could be used to obtain a randomized nonlinear variant of
  LDA.
  
  \paragraph{Extension to RDC.}
  The Randomized Dependence Coefficient or RDC \citep{Lopez-Paz13} is defined as
  the largest canonical correlation of RCCA when performed on the copula
  transformation of the data matrices of $\bm X$ and $\bm Y$. Our analysis
  applies to the further understanding of RDC.

  \section{Experiments} \label{sec:experiments}
  We investigate the performance of RCCA in multiple experiments with
  real-world data against state-of-the-art algorithms. Section \ref{sec:lupi}
  provides a novel algorithm based on RCCA to perform learning using privileged
  information \citep{Vapnik09}. Section \ref{sec:autoencoders} introduces the
  use of RPCA as a tool to train autoencoders in a scalable manner.

  We set our random (Fourier) features to approximate the Gaussian kernel, as
  described in the second paragraph of Section \ref{sec:bochner}. We also
  compare to the Nystr\"om method, set to construct an $m-$dimensional feature
  space formed by the evaluations of the Gaussian kernel on $m$ random points
  from the training set \citep{Yang12}. Gaussian kernel widths $\lbrace
  s_x,s_y\rbrace$ are set using the median heuristic.
 
  \subsection{Empirical Validation of Bernstein
  Inequalities}\label{sec:expbounds} We first turn to the issue of empirically
  validating the bounds obtained in Theorems \ref{thm:pca} and \ref{thm:cca}.
  To do so, we perform simulations in which we separately vary the values of
  the sample size $n$, the number of random projections $m$, and the
  regularization parameter $\gamma$. We use synthetic data matrices $\bm {X}\in
  \mathbb{R}^{n\times 10}$ and $\bm {Y}\in \mathbb{R}^{n\times 10}$ formed by
  i.i.d.\ normal entries. When not varying, the parameters are fixed to
  $n=1000$, $m=1000$ and $\gamma = 10^{-3}$.

  \begin{figure}[h!]
    \begin{center}
    \includegraphics[width=0.5\linewidth]{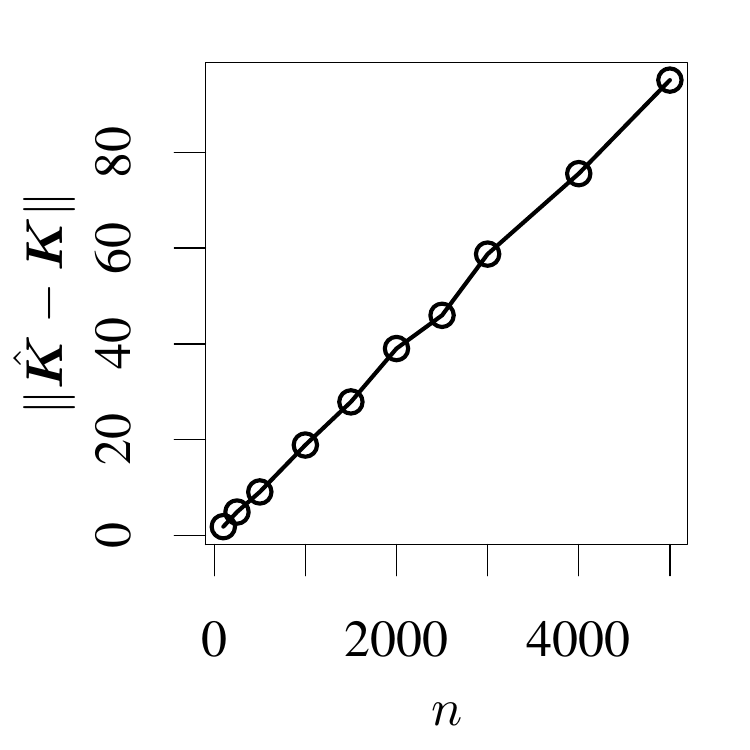}\includegraphics[width=0.5\linewidth]{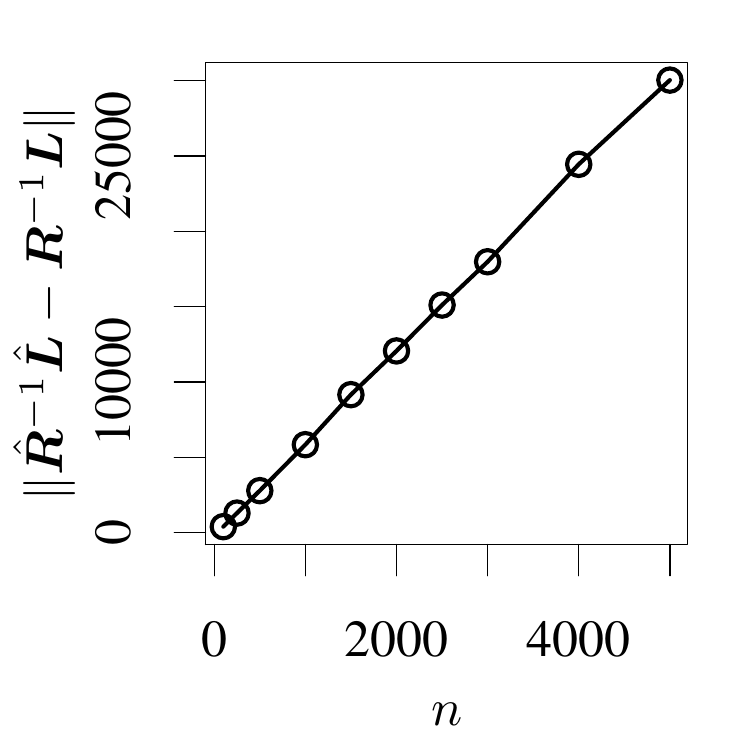}
    \vskip -0.1 cm
    \includegraphics[width=0.5\linewidth]{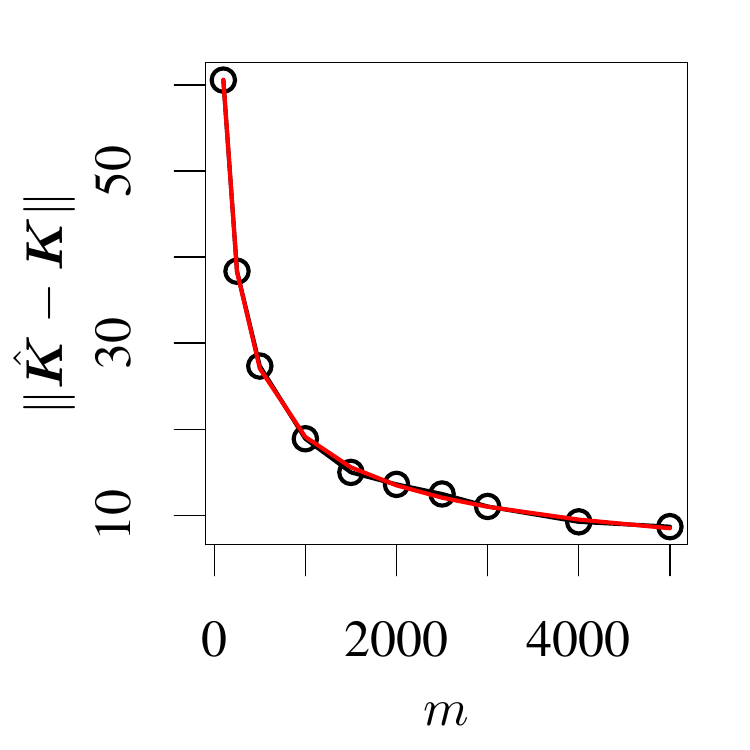}\includegraphics[width=0.5\linewidth]{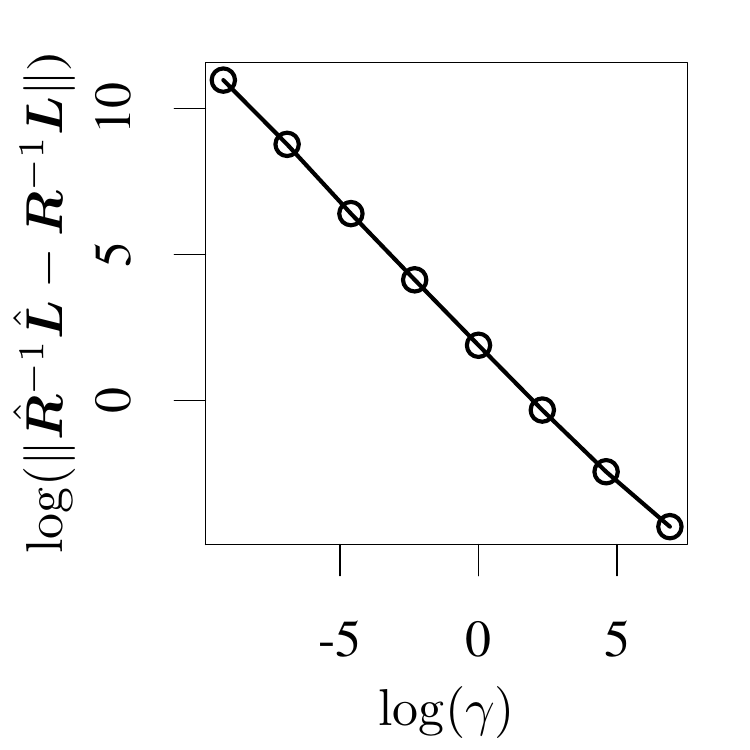}
    \end{center}
    \vskip -0.2 cm
    \caption{Error-norms as a function of a varying parameter, depicted in the
    $x$-axis.  Left: RPCA. Right: RCCA.}
    \label{fig:bounds}
  \end{figure}

  Figure \ref{fig:bounds} depicts the value of the norms from equations
  (\ref{eq:pcaconc}, \ref{eq:ccaconc}) as the parameters $\{n,m,\gamma\}$ vary,
  when averaged over a total of $1000$ random samples $\{\bm X_i, \bm
  Y_i\}_{i=1}^{1000}$. The simulations agree with the presented theoretical
  analysis: the sample size $n$ and regularization parameter $\gamma$ exhibit a
  linear effect, while increasing the number of random features $m$ induces an
  $O(m^{-1/2})$ reduction in error (the closest function $O(m^{-1/2})$ is
  overlaid in red for comparison).
  
  \subsection{Canonical Correlation Analysis}
  We compare three variants of CCA on the task of learning correlated features
  from two modalities of the same data: linear CCA, state-of-the-art Deep CCA
  \citep{Galen13} and the proposed (Fourier and Nystr\"om based) RCCA. We were
  unable to run exact KCCA on the proposed datasets due to its cubic
  complexity; other low-rank approximations such as the one of \citet{Arora12}
  were shown inferior to DCCA, and hence omitted in our analysis. 

  We replicate the two experiments presented in \citet{Galen13}.  The task is
  to measure performance as the accumulated correlation between the canonical
  variables associated with the largest training canonical correlations on some
  unseen test data.  The participating datasets are MNIST and XRMB, which are
  introduced in the following.

  \paragraph{MNIST Handwritten Digits.} Learn correlated representations
  between the left and right halves of the MNIST images \citep{LeCun98}.  Each
  image has a width and height of 28 pixels; therefore, each of the two views
  of CCA consists on 392 features. 54000 random samples are used for training,
  10000 for testing and 6000 to cross-validate the parameters of 
  (D)CCA.

  \paragraph{X-Ray Microbeam Speech Data.} Learn correlated representations of
  simultaneous acoustic and articulatory speech measurements \citep{Westbury94}.
  The articulatory measurements describe the position of the speaker's lips,
  tongue and jaws for seven consecutive frames, yielding a 112-dimensional
  vector at each point in time; the acoustic measurements are 
  the MFCCs for the same frames, producing a
  273-dimensional vector for each point in time. 30000 random samples are used
  for training, 10000 for testing and 10000 to cross-validate the parameters of
  (D)CCA.

  \begin{table}[h!]
  \begin{center}
    \begin{tabular}{|p{1.1cm}|p{1.25cm}|p{1.25cm}|p{1.25cm}|p{1.25cm}|}
      \multicolumn{5}{c}{RCCA on \textbf{MNIST} (50 largest canonical correlations)}\\
      \hline
      \multirow{2}{*}{$m_x,m_y$}  & \multicolumn{2}{|c|}{Fourier} &
      \multicolumn{2}{|c|}{Nystr\"om}\\\cline{2-5}
           & corr. & minutes& corr. & minutes \\\hline
      1000 & 36.31 & 5.55   & 41.68 & 5.29   \\\hline
      2000 & 39.56 & 19.45  & 43.15 & 18.57  \\\hline
      3000 & 40.95 & 41.98  & 43.76 & 41.25  \\\hline
      4000 & 41.65 & 73.80  & 44.12 & 75.00  \\\hline
      5000 & 41.89 & 112.80 & 44.36 & 115.20 \\\hline
      6000 & 42.06 & 153.48 & \textbf{44.49} & 156.07 \\\hline
    \end{tabular}
    \vskip 0.3 cm
    \begin{tabular}{|p{1.1cm}|p{1.25cm}|p{1.25cm}|p{1.25cm}|p{1.25cm}|}
      \multicolumn{5}{c}{RCCA on \textbf{XRMB} (112 largest canonical correlations)}\\
      \hline
      \multirow{2}{*}{$m_x,m_y$}  & \multicolumn{2}{|c|}{Fourier} & \multicolumn{2}{|c|}{Nystr\"om}\\\cline{2-5}
           & corr. & minutes& corr.  & minutes\\\hline
      1000 & 68.79 & 2.95   & 81.82  & 3.07 \\\hline
      2000 & 82.62 & 11.45  & 93.21  & 12.05 \\\hline
      3000 & 89.35 & 26.31  & 98.04  & 26.07 \\\hline
      4000 & 93.69 & 48.89  & 100.97 & 50.07 \\\hline
      5000 & 96.49 & 79.20  & 103.03 & 81.6 \\\hline
      6000 & 98.61 & 120.00 & \textbf{104.47} & 119.4 \\\hline
    \end{tabular}
    \vskip 0.4 cm
    \begin{tabular}{|p{1.1cm}|p{1.25cm}|p{1.25cm}|p{1.25cm}|p{1.25cm}|}
    \cline{2-5}
    \multicolumn{1}{c|}{} & \multicolumn{2}{|c|}{linear CCA} & \multicolumn{2}{|c|}{DCCA} \\\cline{2-5}
    \multicolumn{1}{c|}{} & corr. & minutes & corr. & minutes\\\hline
    \bf MNIST & 28.0 & 0.57 & 39.7 & 787.38  \\\hline
    \bf XRMB  & 16.9 & 0.11 & 92.9 & 4338.32 \\\hline
    \end{tabular}
  \end{center}
    \caption{Sum of largest test canonical correlations and running times by
    all CCA variants in the MNIST and XRMB datasets.}
    \label{table:real}
  \end{table}

  \paragraph{Summary of Results.} Table~\ref{table:real} shows the sum of the
  largest canonical correlations (corr.) obtained by each CCA variant and their
  running times (minutes, single 1.8GHz core) on the MNIST and XRMB test sets.
  Given enough random projections ($m=m_x=m_y$), RCCA is able to explain the most amount of
  test correlation while running drastically faster than
  DCCA\footnote{Running times for DCCA correspond to a single cross-validation
  iteration of its ten hyper-parameters. DCCA has 2 layers for
  MNIST and 8 layers for XRMB.}. Moreover, when using random features (i) the
  number of weights required to be stored at test time for RCCA is up to two
  orders of magnitude lower than for DCCA and (ii) the use of Fastfood
  multiplications \citep{Le13} allows much faster model evaluation.
  
  \paragraph{Parameter Selection.} No parameters were tuned for RCCA: the
  kernel widths were heuristically set and CCA regularization is implicitly 
  provided by the use of randomness (thus set to $10^{-8}$). The number of
  random features $m$ can be set to the maximum value that fits within the
  available (training or test time) computational budget. On the contrary,
  previous state-of-the-art DCCA has ten parameters (two autoencoder parameters
  for pretraining, number of hidden layers, number of hidden units and CCA
    regularizers for each view), which were cross-validated using the grids
    described in \citet{Galen13}.  Cross-validating RCCA parameters did not
    significantly improve performance.

  If desired, further speed improvements for RCCA could be achieved by
  distributing the computation of covariance matrices over several CPUs or
  GPUs, and by making use of truncated SVD routines \citep{BaglamaR06}.

  \subsection{Learning Using Privileged Information}
  \label{sec:lupi}
  In Vapnik's \emph{Learning Using Privileged Information} (LUPI) paradigm
  \citep{Vapnik09} the learner has access to a set of \emph{privileged} features
  or information $\bm X_\star$, exclusive of training time. These features are
  understood as helpful high-level ``teacher explanations'' about each of the
  training samples. The challenge is to build algorithms able to extract
  information from this privileged features at training time in order to
  build a better classifier at test time. We propose to use RCCA to construct a
  highly correlated subspace between the regular features $\bm X$ and the
  privileged features $\bm X_\star$, accessible at test time through a
  nonlinear transformation of $\bm X$.
  
  We experiment with the \emph{Animals-with-Attributes} dataset
  \citep{Lampert09}. In this dataset, the regular features $\bm X$ are the SURF
  descriptors of $30000$ pictures of $35$ different animals; the privileged
  features $\bm X_\star$ are $85$ high-level binary attributes associated with
  each picture (such as \emph{eats-fish} or \emph{can-fly}). To extract
  information from $\bm X_\star$ at training time, we build a feature space
  formed by the concatenation of the $85$, five-dimensional top canonical
  variables $\bm z_x(\bm X)\bm F^{(i)}_{1:5}$ associated with each
  $\mathrm{RCCA}(\bm X, [\bm X_\star^{(i)}, \bm y])$, $i \in \{ 1, \ldots,
  85\}$.  The vector $\bm y$ denotes the training labels.

  \begin{figure}[h!]
    \begin{center}
    \includegraphics[width=\linewidth]{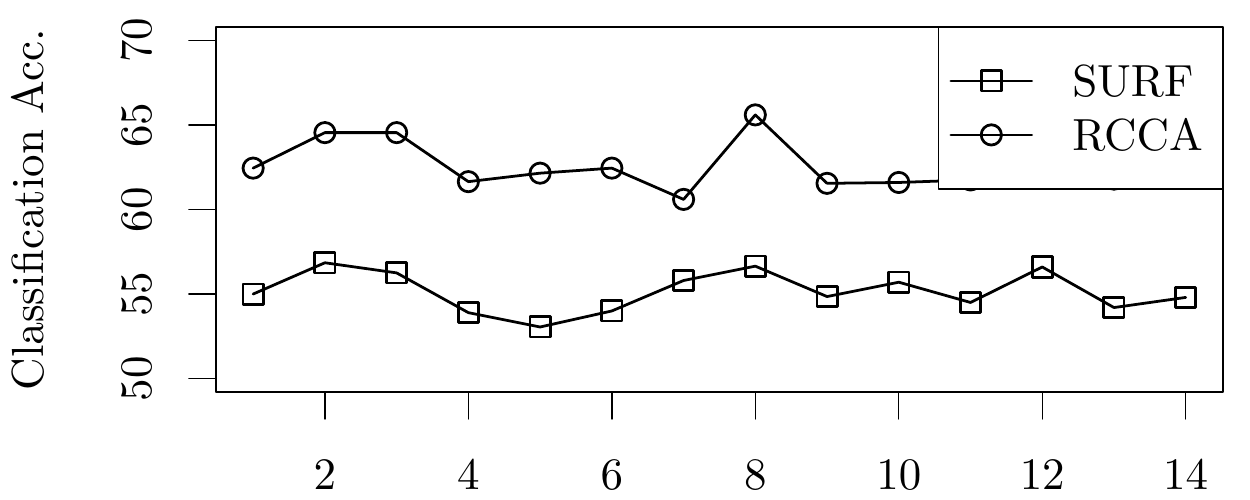}
    \end{center}
    \vskip -0.4 cm
    \caption{Classification accuracy on the LUPI Experiments.}
    \label{fig:animals}
  \end{figure}

  We perform 14 random training/test partitions of $1000$ samples each. Each
  partition groups a random subset of $10$ animals as class ``$0$'' and a
  second random subset of $10$ animals as class ``$1$''. Hence, each experiment 
  is a different, challenging binary classification problem. Figure
  \ref{fig:animals} shows the test classification accuracy of a linear SVM when
  using as features the images' SURF descriptors or the RCCA ``semi-privileged''
  features. As a side note, directly using the high-level attributes yields
  $100\%$ accuracy.  The cost parameter of the linear SVM is cross-validated on
  the grid $[10^{-4}, \ldots, 10^4]$. We observe an average improvement of
  $14\%$ in classification when using the RCCA basis instead of the image
  features alone.  Results are statistically significant respect to a paired
  Wilcoxon test on a $95\%$ confidence interval. The SVM+ algorithm
  \citep{Vapnik09} did not improve on the regular SVM
  using SURF descriptors.

  \subsection{Randomized Autoencoders}\label{sec:autoencoders}
  RPCA can be used for scalable training of nonlinear autoencoders. The process
  involves (i) mapping the observed data $\bm Y\in\mathbb{R}^{D\times n}$ into
  the latent factors $\bm X \in\mathbb{R}^{d\times n}$ using the top $d$
  nonlinear principal components from RPCA and (ii) reconstructing $\bm
  Y$ from $\bm X$ using $D$ nonlinear regressors.
  
  \begin{figure}[h!]
    \begin{center}
    \includegraphics[width=\linewidth]{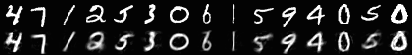}\\
    \includegraphics[width=\linewidth]{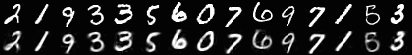}\\
    \includegraphics[width=\linewidth]{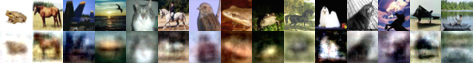}\\
    \includegraphics[width=\linewidth]{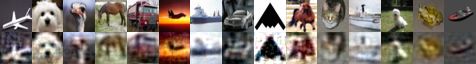}
    \vskip -0.1 cm
    \end{center}
    \vskip -0.2 cm
    \caption{Autoencoder reconstructions of unseen test images for the MNIST (top) and
    CIFAR-10 (bottom) datasets.} 
    \label{fig:autoencoder}
  \end{figure}

  Figure \ref{fig:autoencoder} shows the reconstruction of \emph{unseen} MNIST
  and CIFAR-10 images after being compressed with RPCA.
  The number random projections was set to $m=2000$. The number of latent
  dimensions was set to $d=20$ for MNIST, and $d=40$ (first row) or $d=100$
  (second row) for CIFAR-10.
  Training took under 200 seconds for each full dataset.

  \paragraph{Acknowledgements}
  We thank the anonymous reviewers for their numerous comments, and the
  fruitful discussions had with Yarin Gal, Mark van der Wilk and Maxim
  Rabinovich. Lopez-Paz is supported by Obra Social ``la Caixa''. 

  \clearpage
  \newpage
  \bibliography{rca}

\begin{thebibliography}{34}
\providecommand{\natexlab}[1]{#1}
\providecommand{\url}[1]{\texttt{#1}}
\expandafter\ifx\csname urlstyle\endcsname\relax
  \providecommand{\doi}[1]{doi: #1}\else
  \providecommand{\doi}{doi: \begingroup \urlstyle{rm}\Url}\fi

\bibitem[Achlioptas et~al.(2002)Achlioptas, McSherry, and
  Sch\"olkopf]{achlio02}
Achlioptas, D., McSherry, F., and Sch\"olkopf, B.
\newblock Sampling techniques for kernel methods.
\newblock \emph{NIPS}, 2002.

\bibitem[Andrew et~al.(2013)Andrew, Arora, Livescu, and Bilmes]{Galen13}
Andrew, G., Arora, R., Livescu, K., and Bilmes, J.
\newblock Deep canonical correlation analysis.
\newblock \emph{ICML}, 2013.

\bibitem[Arora \& Livescu(2012)Arora and Livescu]{Arora12}
Arora, R. and Livescu, K.
\newblock Kernel {CCA} for multi-view learning of acoustic features using
  articulatory measurements.
\newblock \emph{MLSLP}, 2012.

\bibitem[Avron et~al.(2013)Avron, Boutsidis, Toledo, and Zouzias]{Avron13}
Avron, H., Boutsidis, C., Toledo, S., and Zouzias, A.
\newblock Efficient dimensionality reduction for canonical correlation
  analysis.
\newblock \emph{ICML}, 2013.

\bibitem[Bach \& Jordan(2002)Bach and Jordan]{Bach02}
Bach, F.~R. and Jordan, M.~I.
\newblock Kernel independent component analysis.
\newblock \emph{JMLR}, 2002.

\bibitem[Baglama \& Reichel(2006)Baglama and Reichel]{BaglamaR06}
Baglama, J. and Reichel, L.
\newblock Restarted block lanczos bidiagonalization methods.
\newblock \emph{Numerical Algorithms}, 2006.

\bibitem[Baldi \& Hornik(1989)Baldi and Hornik]{Baldi89}
Baldi, P. and Hornik, K.
\newblock Neural networks and principal component analysis: Learning from
  examples without local minima.
\newblock \emph{Neural Networks}, 1989.

\bibitem[Bie et~al.(2005)Bie, Cristianini, and Rosipal]{Bie05}
Bie, T.~De, Cristianini, N., and Rosipal, R.
\newblock Eigenproblems in pattern recognition.
\newblock \emph{Handbook of Geometric Computing}, 2005.

\bibitem[Chaudhuri et~al.(2009)Chaudhuri, Kakade, Livescu, and
  Sridharan]{Chaudhuri09}
Chaudhuri, K., Kakade, S.~M., Livescu, K., and Sridharan, K.
\newblock Multi-view clustering via canonical correlation analysis.
\newblock \emph{ICML}, 2009.

\bibitem[Hamid et~al.(2014)Hamid, Xiao, Gittens, and DeCoste]{hamid2013compact}
Hamid, R., Xiao, Y., Gittens, A., and DeCoste, D.
\newblock Compact random feature maps.
\newblock \emph{ICML}, 2014.

\bibitem[Hinton \& Salakhutdinov(2006)Hinton and Salakhutdinov]{Hinton06}
Hinton, G.~E. and Salakhutdinov, R.~R.
\newblock Reducing the dimensionality of data with neural networks.
\newblock \emph{Science}, 2006.

\bibitem[Hotelling(1936)]{Hotelling36}
Hotelling, H.
\newblock {Relations Between Two Sets of Variates}.
\newblock \emph{Biometrika}, 1936.

\bibitem[Jolliffe(2002)]{Jolliffe02}
Jolliffe, I.~T.
\newblock \emph{{Principal Component Analysis}}.
\newblock Springer, 2002.

\bibitem[Kakade \& Foster(2007)Kakade and Foster]{Kakade07}
Kakade, S.~M. and Foster, D.~P.
\newblock Multi-view regression via canonical correlation analysis.
\newblock \emph{COLT}, 2007.

\bibitem[Kar \& Karnick(2012)Kar and Karnick]{kar2012random}
Kar, P. and Karnick, H.
\newblock Random feature maps for dot product kernels.
\newblock \emph{arXiv:1201.6530}, 2012.

\bibitem[Lai \& Fyfe(2000)Lai and Fyfe]{laiFy00}
Lai, P. and Fyfe, C.
\newblock Kernel and nonlinear canonical correlation analysis.
\newblock \emph{International Journal of Neural Systems}, 2000.

\bibitem[Lampert et~al.(2009)Lampert, Nickisch, and Harmeling]{Lampert09}
Lampert, C.~H., Nickisch, H., and Harmeling, S.
\newblock Learning to detect unseen object classes by betweenclass attribute
  transfer.
\newblock \emph{CVPR}, 2009.

\bibitem[Le et~al.(2013)Le, Sarlos, and Smola]{Le13}
Le, Q., Sarlos, T., and Smola, A.
\newblock {Fastfood -- Approximating} kernel expansions in loglinear time.
\newblock \emph{ICML}, 2013.

\bibitem[LeCun \& Cortes(1998)LeCun and Cortes]{LeCun98}
LeCun, Y. and Cortes, C.
\newblock The {MNIST} database of handwritten digits.
\newblock 1998.

\bibitem[Lopez-Paz et~al.(2013)Lopez-Paz, Hennig, and Sch\"olkopf]{Lopez-Paz13}
Lopez-Paz, D., Hennig, P., and Sch\"olkopf, B.
\newblock {The Randomized Dependence Coefficient}.
\newblock \emph{NIPS}, 2013.

\bibitem[Luxburg(2007)]{Luxburg07}
Luxburg, U.
\newblock A tutorial on spectral clustering.
\newblock \emph{Statistics and Computing}, 2007.

\bibitem[{Mackey} et~al.(2014){Mackey}, {Jordan}, {Chen}, {Farrell}, and
  {Tropp}]{Tropp14}
{Mackey}, L., {Jordan}, M.~I., {Chen}, R.~Y., {Farrell}, B., and {Tropp}, J.~A.
\newblock {Matrix Concentration Inequalities via the Method of Exchangeable
  Pairs}.
\newblock \emph{Annals of Probability}, 2014.

\bibitem[Mahoney(2011)]{Mahoney11}
Mahoney, M.~W.
\newblock Randomized algorithms for matrices and data.
\newblock \emph{Foundations and Trends in Machine Learning}, 2011.

\bibitem[McWilliams et~al.(2013)McWilliams, Balduzzi, and
  Buhmann]{McWilliams13}
McWilliams, B., Balduzzi, D., and Buhmann, J.
\newblock Correlated random features for fast semi-supervised learning.
\newblock \emph{NIPS}, 2013.

\bibitem[Pearson(1901)]{Pearson01}
Pearson, K.
\newblock {On lines and planes of closest fit to systems of points in space}.
\newblock \emph{Philosophical Magazine}, 1901.

\bibitem[Rahimi \& Recht(2008)Rahimi and Recht]{Rahimi08}
Rahimi, A. and Recht, B.
\newblock Weighted sums of random kitchen sinks: Replacing minimization with
  randomization in learning.
\newblock \emph{NIPS}, 2008.

\bibitem[Sch\"olkopf \& Smola(2002)Sch\"olkopf and Smola]{Scholkopf01}
Sch\"olkopf, B. and Smola, A.~J.
\newblock \emph{Learning with Kernels: Support Vector Machines, Regularization,
  Optimization, and Beyond}.
\newblock {MIT} {Press}, 2002.

\bibitem[Sch\"olkopf et~al.(1999)Sch\"olkopf, Smola, and
  M\"uller]{Schoelkopf99}
Sch\"olkopf, B., Smola, A., and M\"uller, K.~R.
\newblock Kernel principal component analysis.
\newblock \emph{Advances in kernel methods - Support vector learning}, 1999.

\bibitem[{Tropp}(2012)]{Tropp12b}
{Tropp}, J.~A.
\newblock {User-Friendly Tools for Random Matrices: An Introduction}.
\newblock \emph{NIPS Tutorials}, 2012.

\bibitem[Vapnik \& Vashist(2009)Vapnik and Vashist]{Vapnik09}
Vapnik, V. and Vashist, A.
\newblock A new learning paradigm: Learning using privileged information.
\newblock \emph{Neural Networks}, 2009.

\bibitem[Westbury(1994)]{Westbury94}
Westbury, J.~R.
\newblock {X-Ray} microbeam speech production database user's handbook version
  1.0.
\newblock 1994.

\bibitem[Williams \& Seeger(2001)Williams and Seeger]{Seeger01}
Williams, C. and Seeger, M.
\newblock Using the {Nystr\"om} method to speed up kernel machines.
\newblock \emph{NIPS}, 2001.

\bibitem[Yang et~al.(2014)Yang, Sindhwani, Avron, and Mahoney]{yang14}
Yang, J., Sindhwani, V., Avron, H., and Mahoney, M.~W.
\newblock {Quasi-Monte Carlo Feature Maps for Shift-Invariant Kernels}.
\newblock \emph{ICML}, 2014.

\bibitem[Yang et~al.(2012)Yang, Li, Mahdavi, Jin, and Zhou]{Yang12}
Yang, T., Li, Y., Mahdavi, M., Jin, R., and Zhou, Z.
\newblock Nystr\"{o}m method vs random {Fourier} features: A theoretical and
  empirical comparison.
\newblock \emph{NIPS}, 2012.

\end{thebibliography}
  \bibliographystyle{icml2014}
\end{document}